\documentclass[runningheads]{llncs}
\usepackage{graphicx}

\usepackage{tikz}
\usepackage{comment}
\usepackage{amsmath,amssymb} 
\usepackage{color}
\usepackage{multirow}
\usepackage{makecell}
\usepackage{xspace}
\usepackage{lipsum}
\usepackage[accsupp]{axessibility}  

\usepackage{color, colortbl}
\definecolor{Olive_Green}{rgb}{0.0, 0.55, 0.0}
\definecolor{YellowGreen}{rgb}{0.60, 0.80, 0.20}
\definecolor{light_blue}{RGB}{62,129,183}
\definecolor{diff_color}{RGB}{0, 0, 255}
\newcommand\REV[1]{#1}
\usepackage[pagebackref=false,breaklinks=true,letterpaper=true,colorlinks,  citecolor = Olive_Green, bookmarks=false]{hyperref}
\usepackage{sidecap}
\usepackage{caption}
\usepackage{subcaption}
\usepackage{booktabs}
\usepackage{pifont}
\usepackage{algorithm}
\usepackage{algcompatible}
\usepackage{algpseudocode}
\usepackage{enumitem}

\definecolor{color1}{RGB}{100,178,100}
\definecolor{color2}{RGB}{80,200,180}
\definecolor{color3}{RGB}{60,120,216}
\definecolor{color4}{RGB}{103,78,167}
\definecolor{color5}{RGB}{153,51,102}
\definecolor{color6}{RGB}{166,77,21}
\definecolor{color7}{RGB}{255,102,0}
\definecolor{color8}{RGB}{255,51,204}
\definecolor{color9}{RGB}{153,204,0}
\definecolor{color10}{RGB}{32,4,240}
\definecolor{color11}{RGB}{170,150,70}
\usepackage{multicol}

\newcommand{\IBratio}{$\textcolor{red}{\gamma_R}$}
\newcommand{\IBdepth}{$\textcolor{color5}{\gamma_{N_1}}$}

\newcommand{\Tdima}{$\textcolor{color2}{\gamma_{D_1}}$}
\newcommand{\Tdimb}{$\textcolor{color2}{\gamma_{D_2}}$}
\newcommand{\Tdimc}{$\textcolor{color2}{\gamma_{D_3}}$}
\newcommand{\Tdimd}{$\textcolor{color2}{\gamma_{D_4}}$}
\newcommand{\TdimALL}{$\textcolor{color2}{\gamma_{D_{1-4}}}$}

\newcommand{\Tdimh}{$\textcolor{color3}{\gamma_{E}}$}

\newcommand{\TdimaRaw}{\textcolor{color2}{\gamma_{D_1}}}
\newcommand{\TdimbRaw}{\textcolor{color2}{\gamma_{D_2}}}
\newcommand{\TdimcRaw}{\textcolor{color2}{\gamma_{D_3}}}
\newcommand{\TdimdRaw}{\textcolor{color2}{\gamma_{D_4}}}
\newcommand{\TdimhRaw}{\textcolor{color3}{\gamma_{E}}}

\newcommand{\mlp}{$\textcolor{color7}{\gamma_M}$}

\newcommand{\dtwo}{$\textcolor{color5}{\gamma_{N_2}}$}
\newcommand{\dthree}{$\textcolor{color5}{\gamma_{N_3}}$}
\newcommand{\dfour}{$\textcolor{color5}{\gamma_{N_4}}$}

\newcommand{\dALL}{$\textcolor{color5}{\gamma_{N_{1-4}}}$}

\newcommand{\wstwo}{$\textcolor{color8}{\gamma_{W_2}}$}
\newcommand{\wsthree}{$\textcolor{color8}{\gamma_{W_3}}$}
\newcommand{\wsfour}{$\textcolor{color8}{\gamma_{W_4}}$}
\newcommand{\wsALL}{$\textcolor{color8}{\gamma_{W_{2-4}}}$}

\newcommand{\IBratioRaw}{\textcolor{red}{\gamma_R}}
\newcommand{\IBdepthRaw}{\textcolor{color5}{\gamma_{N_1}}}

\newcommand{\mlpRaw}{\textcolor{color7}{\gamma_M}}
\newcommand{\dtwoRaw}{\textcolor{color5}{\gamma_{N_2}}}
\newcommand{\dthreeRaw}{\textcolor{color5}{\gamma_{N_3}}}
\newcommand{\dfourRaw}{\textcolor{color5}{\gamma_{N_4}}}
\newcommand{\wstwoRaw}{\textcolor{color8}{\gamma_{W_2}}}
\newcommand{\wsthreeRaw}{\textcolor{color8}{\gamma_{W_3}}}
\newcommand{\wsfourRaw}{\textcolor{color8}{\gamma_{W_4}}}

\def \alambic {\includegraphics[width=0.02\linewidth]{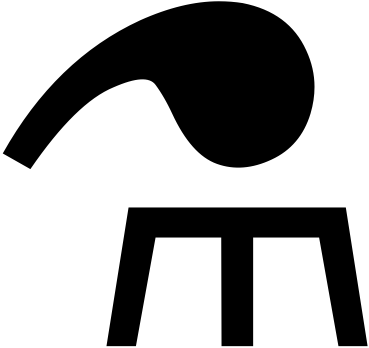}\xspace}

\newcommand\blfootnote[1]{%
  \begingroup
  \renewcommand\thefootnote{}\footnote{#1}%
  \addtocounter{footnote}{-1}%
  \endgroup
}

\begin{document}
\pagestyle{headings}
\mainmatter
\def\ECCVSubNumber{4537}  

\title{TinyViT: Fast Pretraining Distillation for \\  Small Vision Transformers}

\titlerunning{TinyViT}
\author{Kan Wu$^{1, *}$, Jinnian Zhang$^{2, *}$, Houwen Peng$^{1, *, \dagger}$,  \\ Mengchen Liu$^{2}$,
Bin Xiao$^{2}$, Jianlong Fu$^{1}$, Lu Yuan$^{2}$\\
}
\authorrunning{Kan Wu et al.}

\institute{$^1$ Microsoft Research,  $^2$ Microsoft Cloud+AI
}

\maketitle

\blfootnote{
  $^*$Equal contribution. Work done when Kan and Jinnian were interns of Microsoft.
  \\
  ~$^\dagger$Corresponding author: houwen.peng@microsoft.com.
}

\begin{abstract}
Vision transformer (ViT) recently has drawn great attention in computer vision due to its remarkable model capability. However, most prevailing ViT models suffer from huge number of parameters, restricting their applicability on devices with limited resources. To alleviate this issue, we propose TinyViT, a new family of tiny and efficient small vision transformers pretrained on large-scale datasets with our proposed fast distillation framework.
The central idea is to transfer knowledge from large pretrained models to small ones, while enabling small models to get the dividends of massive pretraining data. More specifically, we apply distillation during pretraining  for knowledge transfer. The logits of large teacher models are sparsified and stored in disk in advance to save the memory cost and computation overheads. The tiny student transformers are automatically scaled down from a large pretrained model with 
computation and parameter constraints.
Comprehensive experiments demonstrate the efficacy of TinyViT. It achieves a top-1 accuracy of 84.8\% on ImageNet-1k with only 21M parameters, being comparable to Swin-B pretrained on ImageNet-21k while using 4.2 times fewer parameters.
Moreover, increasing image resolutions, TinyViT can reach 86.5\% accuracy, being slightly better than Swin-L while using only 11\% parameters. 
Last but not the least, we demonstrate a good transfer ability of TinyViT on various downstream tasks. Code and models are available at \href{https://github.com/microsoft/Cream/tree/main/TinyViT}{here}.

\end{abstract}

\section{Introduction}

Transformer \cite{vaswani2017attention} has taken computer vision domain by storm and are becoming increasingly popular in both research and practice \cite{ViT,detr,scalevit}. One of the recent trends for vision transforms (ViT) is to continue to grow in model size while yielding improved performance on standard benchmarks \cite{scalevit,swin_v2,vmoe}. For example, V-MoE~\cite{vmoe} uses 305 million images to train an extremely large model with 14.7 billion parameters, achieving state-of-the-art performance on image classification. Meanwhile, the Swin transformer uses 3 billion parameters with 70 million pretraining images, to attain promising results on downstream detection and segmentation tasks~\cite{swin,swin_v2}. 
Such large model sizes and the accompanying heavy pretraining costs  make these models unsuitable for applications involving limited computational budgets, such as mobile and IoT edge devices~\cite{MiniViT}.

\begin{figure}[t]

	\centering
	\begin{minipage}[c]{.4\textwidth}
		\begin{flushleft}
			\caption{\footnotesize Comparison of our TinyViT with other small vision transformer models~\cite{deit,swin} on ImageNet-1k in terms of w/ and w/o ImageNet-21k pretraining and distillation. Pretraining with distillation can effectively improve the performance of all these small transformer models, further unveiling their capacities. Best viewed in color. }\label{fig:motivation}
		\end{flushleft}
	\end{minipage}%
	\begin{minipage}[c]{.6\textwidth}
		\centering
		\begin{flushright}
			\includegraphics[width=1.0\textwidth,trim=10 160 20 160]{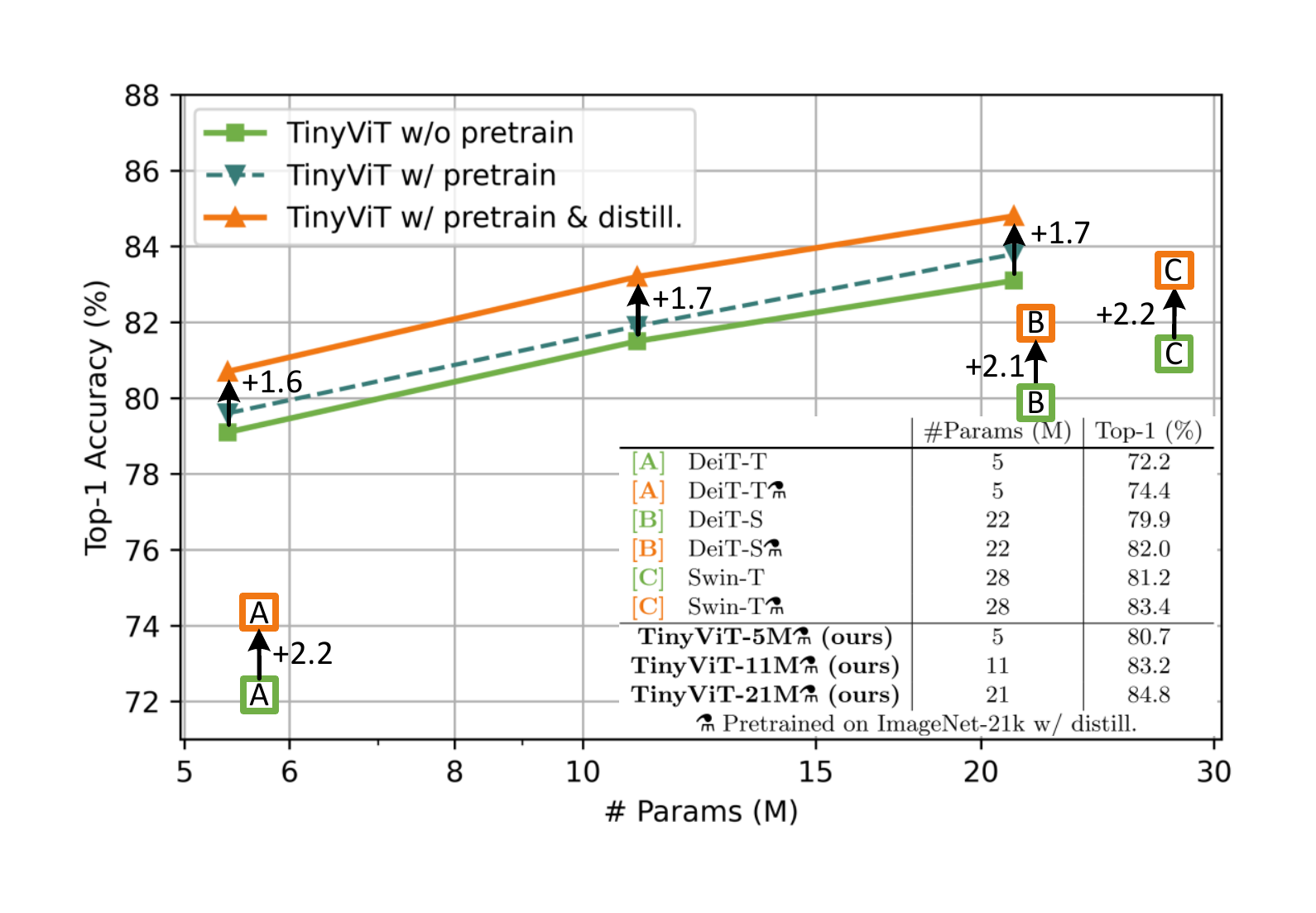}
		\end{flushright}
	\end{minipage}
	
	\vspace{-6mm}
\end{figure}

In contrast to scaling up models to large scales, this work turns attention to downsizing vision transformers, aiming to generate a new family of tiny models and elevate their transfer capacities in downstream tasks. In particular, we explore the following key issue: \textit{how to effectively transfer the knowledge of existing large-scale transformers to small ones, as well as unleash the power of large-scale data to elevate the representability of small models?} In computer vision, it has long been recognized that large models pretrained on large datasets often achieve better results, while small models easily become saturated (or underfitting) as the growth of data \cite{scalevit,swin_v2}.  
Is there any possible way for small models to absorb knowledge from massive data and further unveil their capacities?

To answer this question, we introduce a fast knowledge distillation method to pretrain small models, and show that small models can also get the dividends of massive pretraining data with the guidance of large models. 
More specifically, we observe that direct pretraining of small models suffers from performance saturation, especially when the data scale increases.
But if we impose distillation during pretraining, using a powerful model as the teacher, the potentials of large-scale pretraining data can be unlocked for small models, as demonstrated in Fig.~\ref{fig:motivation}.
Meanwhile, the distilled small models can be transferred well to downstream tasks, since they have learned a great deal of knowledge about how to generalize from the large model as well as the large-scale pretraining data. We give a detailed discussion in Sec.~\ref{sec:analysis} exploring the underlying reasons why pretraining distillation is able to further unveil the capacities of small models. 

Pretraining models with distillation is inefficient and costly, because a considerable proportion of computing resources is consumed on passing training data through the large teacher model in each iteration, rather than training the target small student. Also, a giant teacher may occupy the most GPU memory, significantly slowing down the training speed of the students (due to limited batch size). To address this issue, we propose a fast and scalable distillation strategy. More concretely, we propose to generate a sparse probability vector as the soft label of each input image in advance, and store it into label files together with the corresponding data augmentation information like random cropping, RandAugment~\cite{randaug}, CutMix~\cite{cutmix}, \emph{etc}. During training, we reuse the stored sparse soft labels and augmentations to precisely replicate the distillation procedure, successfully omitting the forward computation and storage of large teacher models. Such strategy has two advantages: 1) Fast. It largely saves the memory cost and computation overheads of generating teachers' soft labels during training. Thus, the distillation of small models can be largely speed up because it is able to use much larger batch size.
\REV{Besides, since the teacher logits per epoch are independent, they can be saved in parallel, instead of epoch-by-epoch in conventional methods.} 2) Scalable. It can mimic any kind of data augmentation and generate the corresponding soft labels. We just need to forward the large teacher model for only once, and reuse the soft labels for arbitrary student models.

We verify the efficacy of our fast pretraining distillation framework not only on existing small vision transformers, such as DeiT-T \cite{deit} and Swin-T \cite{swin}, but also over our new designed tiny architectures. Specifically, following \cite{x3d}, we adopt a progressive model contraction approach to scale down a large model and generate a family of tiny vision transformers (TinyViT). With our fast pretraining distillation on ImageNet-21k~\cite{imagenet}, TinyViT with 21M parameters achieves 84.8\% top-1 accuracy on ImageNet-1k, being 4.2 times smaller than the pretrained Swin-B (85.2\% accuracy with 88M parameters). With higher resolution, our model can reach 86.5\% top-1 accuracy, establishing new state-of-the-art performance on ImageNet-1k under aligned settings. Moreover, TinyViT models demonstrate good transfer capacities on downstream tasks. For instance, TinyViT-21M gets an AP of 50.2 on COCO object detection benchmark, being 2.1 points superior to Swin-T using 28M parameters.

In summary, the main cotributions of this work are twofold.
\vspace{-1mm}
\begin{itemize}
    \item We propose a fast pretraining distillation framework to unleash the capacity of small models by fully leveraging the large-scale pretraining data. To our best knowledge, this is the first work exploring small model pretraining. 
    
    \item We release a new family of tiny vision transformer models, which strike a good trade-off between computation and accuracy. With pretraining distillation, such models demonstrate good transfer ability on downstream tasks.
\end{itemize}

\vspace{-5mm}
\section{Related Work}
\vspace{-2mm}

In this section, we review the related work on large-scale pretraining, small vision transformers, and knowledge distillation. 
It is notable that our work is orthogonal to existing literature on model compression techniques such as quantization~\cite{ptq,jacob2018quantization,han2015deep,ptq} and pruning~\cite{VTP,NViT,uvc,spvit}.
These techniques can be used as a post-processing for our TinyViT to further improve model efficiency.

\textbf{Large-scale pretraining.} Bommasani \emph{et al}.~\cite{foundation} first coined the concept of foundation models that are pretrained from large-scale data and have outstanding performance in various downstream tasks. 
For example, BERT~\cite{bert} and GPT-3~\cite{gpt} have been demonstrated to be effective foundation models in natural language processing.
Recently, there are some research efforts in developing foundation models in computer vision, including CLIP~\cite{clip}, Align \cite{align} and Florence~\cite{florence}.
They have shown impressive transfer and zero-shot capabilities. However, these large models are unsuitable for downstream applications with limited computational budgets.
By contrast, our work investigates the pretraining method for small models and improves their transferability to various downstream tasks.

\textbf{Small vision transformers.} 
Lightweight CNNs have powered many mobile vision tasks \cite{mobilenetv3,efficientnet}. 
Recently, there are several attempts developing light vision transformers (ViTs).
Mehta \emph{et al}.~\cite{mobilevit} combined standard convolutions and transformers to develop MobileViT, which outperforms the prevailing MobileNets \cite{mobilenetv3} and ShuffleNet \cite{shufflenet}.  
Gong \emph{et al.}~\cite{nasvit} employed NAS and identified a family of efficient ViTs with MACs ranging from 200M to 800M, surpassing the state-of-the-art. 
Graham \emph{et al.}~\cite{levit} optimized the inference time of small and medium-sized ViTs and generated a family of throughput-efficient ViTs. 
Different from these manually designed or automatically searched small models, our work explores model contraction to generate small models by progressively slimming a large seed model, which can be considered as a complementary work to existing literature on scaling-up large vision transformers~\cite{autoscaling,scalevit,vmoe,swin_v2}.

\textbf{Knowledge distillation.} 
Distillation in a teacher-student framework~\cite{distill_hinton} is widely used to 
leverage knowledge from large teacher models.
It has been extensively studied in convolutional networks~\cite{gou2021knowledge}.
Recently, there are several research works in developing distillation techniques for ViTs~\cite{uvc,meta}.
For example, Touvron et al.~\cite{deit} introduced a distillation token to allow the transformer to learn from a ConvNet teacher, while Jia et al.~\cite{jia2021efficient} proposed to excavate knowledge from the teacher transformer via the connection between images and patches. Distillation for ViTs is still under-explored, especially for pretraining distillation. 

In knowledge distillation, the mostly related work to ours is the recent FKD~\cite{fkd}. 
Both methods share a similar spirit on saving teacher logits to promote training efficiency, but our framework has two advantages. 
1) More efficient. 
Instead of saving the explicit information of each transformation in data augmentation using hundreds of bytes, such as crop coordinates and rotation degree, our framework only needs 4 bytes to store a random seed. The seed will be used as the initial state of the random number generator to reproduce the number sequence that controls the transformations in data augmentation to generate crop coordinates and rotation degree, \emph{etc}.
2) More general. Our framework supports all existing types of data augmentation including the complex Mixup \cite{mixup} and Cutmix \cite{cutmix}, which are not explored in FKD. Moreover, the studied problem in \cite{fkd} is different to ours. We focus on pretraining-stage distillation for transformers, while FKD explores finetune-stage distillation for CNN models. 

\begin{figure*}[t]
  \centering
\includegraphics[width=1.0\linewidth, trim=20 24 30 32]{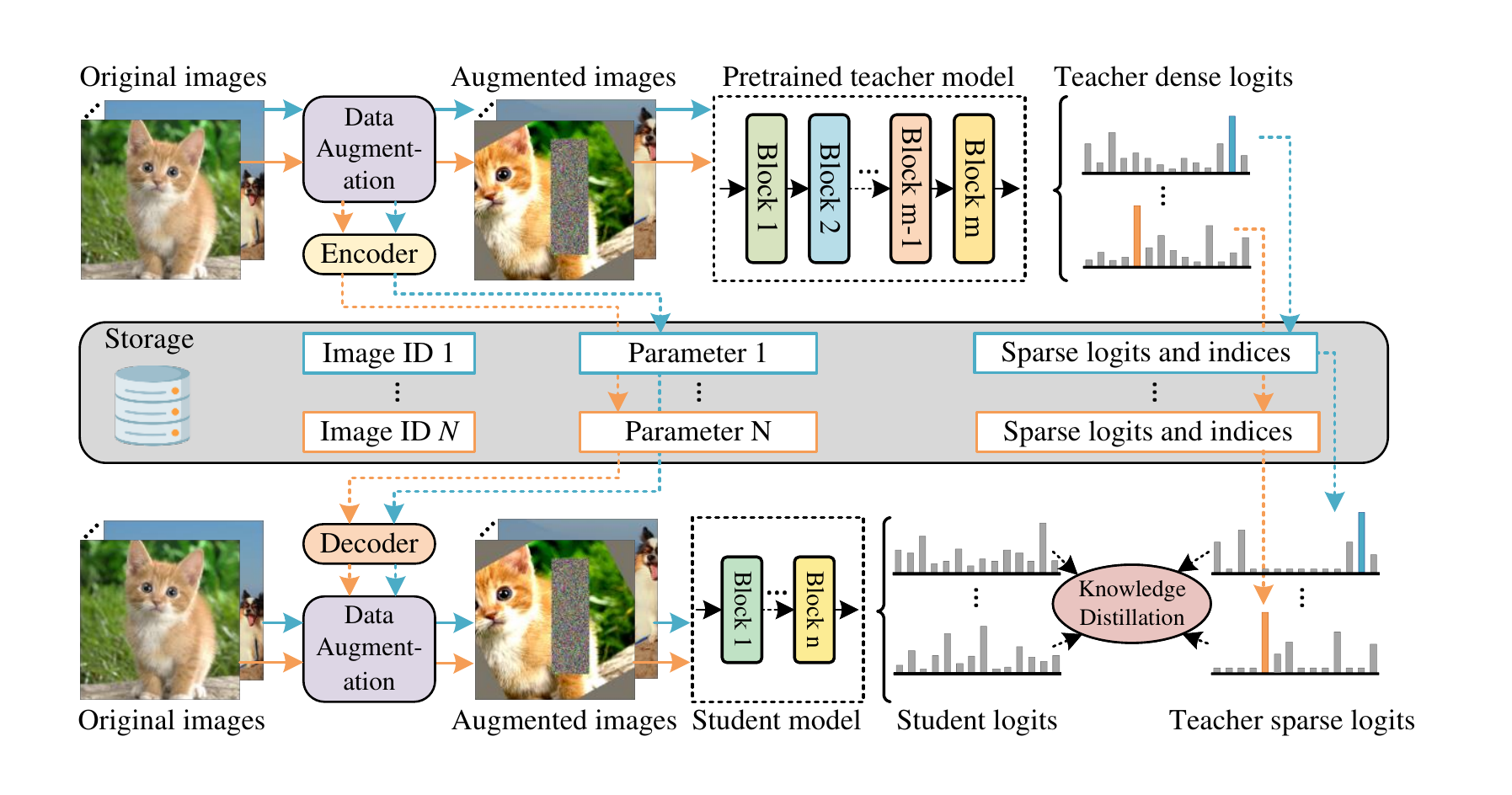}
  \caption{Our fast pretraining distillation framework. \textbf{Top: }the branch for saving teacher logits. Encoded data augmentation and sparsified teacher logits are saved. \textbf{Middle: }the disk for storing information. \textbf{Bottom: }the branch for training the student. The decoder reconstructs the data augmentation, and distillation is conducted between the teacher logits and student outputs. Note that the two branches are independent and asynchronous.}
  \label{fig:framework}
  \vspace{-4mm}
\end{figure*}

\vspace{-5mm}
\section{TinyViT}
\vspace{-2mm}

This section proposes TinyViT, a new family of tiny and efficient models with fast pretraining distillation on large-scale data. We first introduce the fast knowledge distillation framework for small model pretraining in Sec.~\ref{sec: pd}. Then we design a new tiny model family with good computation/accuracy trade-off by progressivley scaling down a large seed model in Sec.~\ref{sec: dsp}.

\vspace{-4mm}
\subsection{Fast Pretraining Distillation}
\vspace{-2mm}
\label{sec: pd}

We observe that direct pretraining of small models  on  massive  data  does  not bring much gains, especially when transferring them to downstream tasks, as presented in Fig.~\ref{fig:motivation}. To address this issue, we resort to knowledge distillation to further unveil the power of pretraining for small models. 
Different from prior work that pays most attention to finetune-stage distillation 
\cite{deit}, we focus on pretraining distillation, which not only allows small models to learn from large-scale model, but also elevates their transfer capacities for downstream tasks.

Pretraining with distillation is inefficient and costly, because a considerable proportion of computing resources is consumed on passing training data through the large teacher model in each iteration, rather than training the target small student. Also, a giant teacher may occupy the most GPU memory, slowing down the training speed of the target students (due to limited batch size). To solve this problem, we propose a fast pretraining distillation framework. As depicted in Fig.~\ref{fig:framework}, we store the information of data augmentation and teacher predictions in advance. During training, we reuse the stored information to precisely replicate the distillation procedure, successfully omitting the forward computation and \REV{memory occupation} of the large teacher model.

\definecolor{fig_orange}{RGB}{245,157,86}
\definecolor{fig_blue}{RGB}{75,172,198}

Mathematically, for an input image $x$ with strong data augmentation $\mathcal{A}$, such as RandAugment~\cite{randaug} and CutMix~\cite{cutmix}, we store both $\mathcal{A}$ and teacher prediction $\hat{\mathbf{y}}=T(\mathcal{A}(x))$, where $T(\cdot)$ and $\mathcal{A}(x)$ are the teacher model and the augmented image. It is notable that passing the same image through the same data augmentation pipeline multiple times will generate different augmented images due to the inherent randomness in data augmentation. Therefore, the pair ($\mathcal{A}$, $\hat{\mathbf{y}}$) needs to be saved for each image in each iteration, as illustrated in Fig.~\ref{fig:framework}.

In the training process, we only need to recover the pairs ($\mathcal{A}$, $\hat{\mathbf{y}}$) from stored files, and optimize the following objective function for student model distillation: 

\vspace{-4mm}
\begin{align}
    \mathcal{L}= CE\left(\hat{\mathbf{y}}, S(\mathcal{A}(x))\right),
    \label{distill_ours}
\end{align}
where $S(\cdot)$ and $CE(\cdot)$ are the student model and cross entropy loss, respectively.
Note that our framework is label-free, \emph{i.e.}, with no need for ground-truth labels, because we only use the soft labels generated by teacher models for training. 
Therefore, it can utilize numerous off-the-shelf web data without labels for large-scale pretraining. Such a label-free strategy is workable in practice because the soft labels are accurate enough while carrying a lot of discriminative information for classification such as category relations.
\REV{We also observe that distillation with ground-truth would cause slight performance drops. The reason may be that not all the labels in ImageNet-21k~\cite{imagenet} are mutually exclusive~\cite{ridnik2021imagenet}, including correlative pairs like ``chair'' and ``furniture'', ``horse'' and ``animal''. Therefore, the one-hot ground-truth label could not describe an object precisely, and in some cases it suppresses either child classes or parent classes during training.}
Moreover, our distillation framework is as fast as training models without distillation since the cumbersome teacher $T(\cdot)$ is removed during training in Eq.~(\ref{distill_ours}).

Besides, our distillation framework is fast due to two key components: sparse soft labels and data augmentation encoding. They can largely reduce the storage consumption while improving memory efficiency during training.

\textbf{Sparse soft labels.}
Let's consider the teacher model outputs $C$ logits for the prediction. It often consumes much storage space to save the whole dense logits of all augmented images if $C$ is large, \emph{e.g.}, $C=21,841$ for ImageNet-21k. Therefore,
we just save the most important part of the logits, \emph{i.e.}, sparse soft labels. Formally, we select the top-$K$ values in $\hat{\mathbf{y}}$, \emph{i.e.},  $ {\{\hat{y}_{\mathcal{I}(k)}\}}_{k=1}^{K} \in \hat{\mathbf{y}}$, and store them along with their indices $\{\mathcal{I}(k)\}_{k=1}^K$ into our label files. During training, we only reuse the stored sparse labels for distillation with label smoothing~\cite{labelsmooth,labelsmooth_zhiqiang}, which is defined as
\vspace{-2mm}
\begin{equation}
    \hat{y}_c =
    \begin{cases}
    ~\hat{y}_{\mathcal{I}(k)} & \text{if } c=\mathcal{I}(k), \\
    ~\frac{1-\sum_{k=1}^K{\hat{y}_{\mathcal{I}(k)}}}{C-K} & \text{otherwise,}
    \end{cases}
\end{equation}
where $\hat{y}_c$ is the recovered teacher logits for student model distillation, \emph{i.e.}, $\hat{\textbf{y}}=[\hat{y}_1,\dots,\hat{y}_c,\dots,\hat{y}_C]$. When the sparsity factor $K$ is small, \emph{i.e.} $K\ll C$, it can reduce logits' storage by orders of magnitude. Moreover, we empirically show that such sparse labels can achieve comparable performance to the dense labels for knowledge distillation, as presented in Sec.~\ref{sec:ablation}.

\textbf{Data augmentation encoding.}
Data augmentation involves a set of parameters $\mathbf{d}$, such as the rotation degree and crop coordinates, to transform the input image. Since $\mathbf{d}$ is different for each image in each iteration, saving it directly becomes memory-inefficient. To solve this problem, we encode $\mathbf{d}$ by a single parameter $d_0=\mathcal{E}(\mathbf{d})$, where $\mathcal{E}(\cdot)$ is the encoder in Fig.~\ref{fig:framework}. Then in the training process, we recover $\mathbf{d}=\mathcal{E}^{-1}(d_0)$ after loading $d_0$ in the storage files, where $\mathcal{E}^{-1}(\cdot)$ is viewed as the decoder. Therefore, the data augmentation can be accurately reconstructed. In practice, a common choice for the decoder is the pseudo-random number generator (i.e. PCG~\cite{pcg}). It takes a single parameter as the input and generates a sequence of parameters. As for the encoder, we simply implement it by a generator for $d_0$ and reusing the decoder $\mathcal{E}^{-1}(\cdot)$. It outputs $\mathbf{d}=\mathcal{E}^{-1}(d_0)$ for the teacher model. $d_0$ is saved for the decoder to reproduce $\mathbf{d}$ when training the student. Thus, the implementation becomes more efficient.

\vspace{-4mm}
\subsection{Model Architectures}
\label{sec: dsp}

In this subsection, we present a new family of tiny vision transformers by scaling down a large model seed with a  progressive model contraction approach \cite{x3d}. 
Specifically, we start with a large model and define a basic set of contraction factors. Then in each step, smaller candidate models are generated around the current model by adjusting the contraction factors.  We select models that satisfy both constraints on the number of parameters and throughput. The model with \REV{the best validation accuracy} will be utilized for further reduction in the next step until the target is achieved. This is a form of \textit{constrained local search}~\cite{localsearch} in the model space spanned by the contraction factors. 

\REV{We adopt a hierarchical vision transformer as the basic architecture, for the convenience of dense prediction downstream tasks like detection that require multi-scale features. More concretely, our base model consists of four stages with a gradual reduction in resolution similar to Swin~\cite{swin} and LeViT~\cite{levit}. The patch embedding block consists of two convolutions with kernel size 3, stride 2 and padding 1. We apply lightweight and efficient MBConvs~\cite{mobilenetv3} in Stage 1 and down sampling blocks, since convolutions at earlier layers are capable of learning low-level representation efficiently due to their strong inductive biases~\cite{levit,early_conv}. The last three stages are constructed by transformer blocks, with window attention to reduce computational cost. The attention biases~\cite{levit} and a $3\times3$ depthwise convolution between attention and MLP are introduced to capture local information~\cite{irpe,chu2021conditional}. Residual connection~\cite{resnet} is applied on each block in Stage 1, as well as attention blocks and MLP blocks. All activation functions are GELU~\cite{gelu}. The normalization layers of convolution and linear are BatchNorm~\cite{BN} and LayerNorm~\cite{LN}, respectively.}

\textbf{Contraction factors.} 
We consider the following factors to form a model:
\vspace{-1mm}
\begin{itemize}
    \item {\TdimALL}: embeded dimension of four stages respectively. Decreasing them results in a thinner network with fewer heads in multi-head self-attention. 
    \item {\dALL}: the number of blocks in four stages respectively. The depth of the model is decreased by reducing these values.
    \item {\wsALL}: window size in the last three stages respectively. As these values become smaller, the model has fewer parameters and higher throughput.

    \item {\IBratio}: channel expansion ratio of the MBConv block. We can obtain a smaller model size by reducing this factor.
    \item {\mlp}: expansion ratio of MLP for all transformer blocks. The hidden dimension of MLP will be smaller if scaling down this value.
    \item {\Tdimh}: \REV{the dimension of each head in multi-head attention. The number of heads will be increased when scaling it down, bringing lower computation cost. }
\end{itemize}

\REV{We scale down the above factors with a progressive model contraction approach~\cite{x3d} and generate a new family of tiny vision transformers: All models share the same factors: \{\IBdepth, \dtwo, \dthree, \dfour\} = \{2, 2, 6, 2\}, \{\wstwo, \wsthree, \wsfour\} = \{7, 14, 7\} and \{\IBratio, \mlp, \Tdimh\} = \{4, 4, 32\}. For the embeded dimensions \{\Tdima, \Tdimb, \Tdimc, \Tdimd\}, TinyViT-21M: \{96, 192, 384, 576\}, TinyViT-11M: \{64, 128, 256, 448\} and TinyViT-5M: \{64, 128, 160, 320\}.}

\vspace{-3mm}
\section{Analysis and Discussions}
\label{sec:analysis}

In this section, we provide analysis and discussions on two key questions: 1) What are the underlying factors limiting small models to fit large data? 2) Why distillation can unlock the power of large data for small models? 
To answer the above questions, we conduct experiments on the widely used large-scale benchmark ImageNet-21k~\cite{imagenet}, which contains 14M images with 21,841 categories

\textit{What are the underlying factors limiting small models to fit large data?} We observe that there are many hard samples existing in IN-21k, \emph{e.g.}, images with wrong labels and similar images with different labels due to the existence of multiple equally prominent objects in the images. This is also recognized by existing literature~\cite{relabel,imagenet_real,ridnik2021imagenet} and approximately 10\% images in ImageNet are considered as hard samples. Small models struggle to fit these hard samples, leading to low training accuracy compared to large models (TinyViT-21M: 53.2\% \emph{vs.} Swin-L-197M~\cite{swin}: 57.1\%) and limited transferability on ImageNet-1k
(TinyViT-21M w/ pretraining: 83.8\% \emph{vs.} w/o pretraining: 83.1\%).

To verify the impact of hard samples, we resort to two techniques. 1) Inspired by \cite{imagenet_real}, we exploit the powerful pretrained model Florence~\cite{florence} finetuned on ImageNet-21k to identify the images whose labels lie outside the top-5 predictions of Florence. Through this procedure, we remove 2M images from ImageNet-21k, approximately 14\%, and then pretrain TinyViT-21M and Swin-T on the cleaned dataset. 
2) We perform distillation to pretrain TinyViT-21M/Swin-T using Florence as the teacher model, which generates soft labels to replace the polluted groundtruth labels in ImageNet-21k. The results of the pretrained models with finetuning on ImageNet-1k are reported in Tab.~\ref{tab:hard_samples}.

We obtain several insights from the results. 1) Pretraining small models on the original ImageNet-21k dataset brings limited performance gains on ImageNet-1k (0.7\% for both Swin-T and TinyViT-21M). 2) After removing parts of the hard samples in ImageNet-21k, both models can better leverage the large data and achieve higher performance gains (1.0\%/1.1\% for Swin-T/TinyViT-21M). 3) Distillation is able to avoid the defects of hard samples, because it does not use the groundtruth labels that are the main cause of hard samples. Thus, it gets higher improvements (2.2\%/1.7\% for Swin-T and TinyViT-21M).

\begin{table}[t]
\begin{center}
\caption{Impact of hard samples. Models are pretrained on IN-21k and then finetuned on IN-1k.
}
\label{tab:hard_samples}
\setlength{\tabcolsep}{6pt}{
\scalebox{0.8}{
\begin{tabular}{c|c|c|ccc}
\Xhline{2\arrayrulewidth}
\multirow{2}{*}{\#} & \multirow{2}{*}{Model} & Pretraining &IN-1k & IN-Real~\cite{imagenet_real} & IN-V2~\cite{imagenet_v2}\\
& &  Dataset & Top-1(\%)&Top-1(\%) &Top-1(\%) \\
\Xhline{2\arrayrulewidth}

 0 & \multirow{4}{*}{Swin-T~\cite{swin}}& Train from scratch on IN-1k & 81.2 & 86.7 & 69.7 \\ %
 \cline{3-6}
  1 & & Original IN-21k & 81.9{\textcolor{YellowGreen}{\scriptsize{(+0.7)}}} &87.0{\textcolor{YellowGreen}{\scriptsize{(+0.3)}}} & 70.6{\textcolor{YellowGreen}{\scriptsize{(+0.9)}}} \\ %
  2 & & Cleaned IN-21k & 82.2{\textcolor{YellowGreen}{\scriptsize{(+1.0)}}} & 87.3{\textcolor{YellowGreen}{\scriptsize{(+0.6)}}} & 71.1{\textcolor{YellowGreen}{\scriptsize{(+1.4)}}} \\ %
  3 & & Original IN-21k w/ distillation & 83.4{\textcolor{YellowGreen}{\scriptsize{(+2.2)}}} & 88.0{\textcolor{YellowGreen}{\scriptsize{(+1.3)}}} & 72.6{\textcolor{YellowGreen}{\scriptsize{(+2.9)}}} \\ %
  \hline
   4 & \multirow{4}{*}{TinyViT-21M  (\textbf{ours})}& Train from scratch on IN-1k & 83.1 & 88.1 & 73.1 \\ %
   \cline{3-6}
  5 & & Original IN-21k & {83.8\textcolor{YellowGreen}{\scriptsize{(+0.7)}}} &{88.4\textcolor{YellowGreen}{\scriptsize{(+0.3)}}} & {73.8\textcolor{YellowGreen}{\scriptsize{(+0.7)}}}  \\ %
    6 & & Cleaned IN-21k & {84.2\textcolor{YellowGreen}{\scriptsize{(+1.1)}}} & {88.5\textcolor{YellowGreen}{\scriptsize{(+0.4)}}} & {73.8\textcolor{YellowGreen}{\scriptsize{(+0.7)}}} \\ %
  7 & & Original IN-21k w/ distillation & {84.8\textcolor{YellowGreen}{\scriptsize{(+1.7)}}} & {88.9\textcolor{YellowGreen}{\scriptsize{(+0.8)}}} & {75.1\textcolor{YellowGreen}{\scriptsize{(+2.0)}}} \\ %
 \Xhline{2\arrayrulewidth}
\end{tabular}}}
\end{center}
\vspace{-4mm}
\end{table}

\begin{figure*}[t]
    \begin{minipage}[b]{1.0\textwidth}
    \subfloat[Teacher]{
        \includegraphics[height=2.9cm, trim=20 50 20 50]{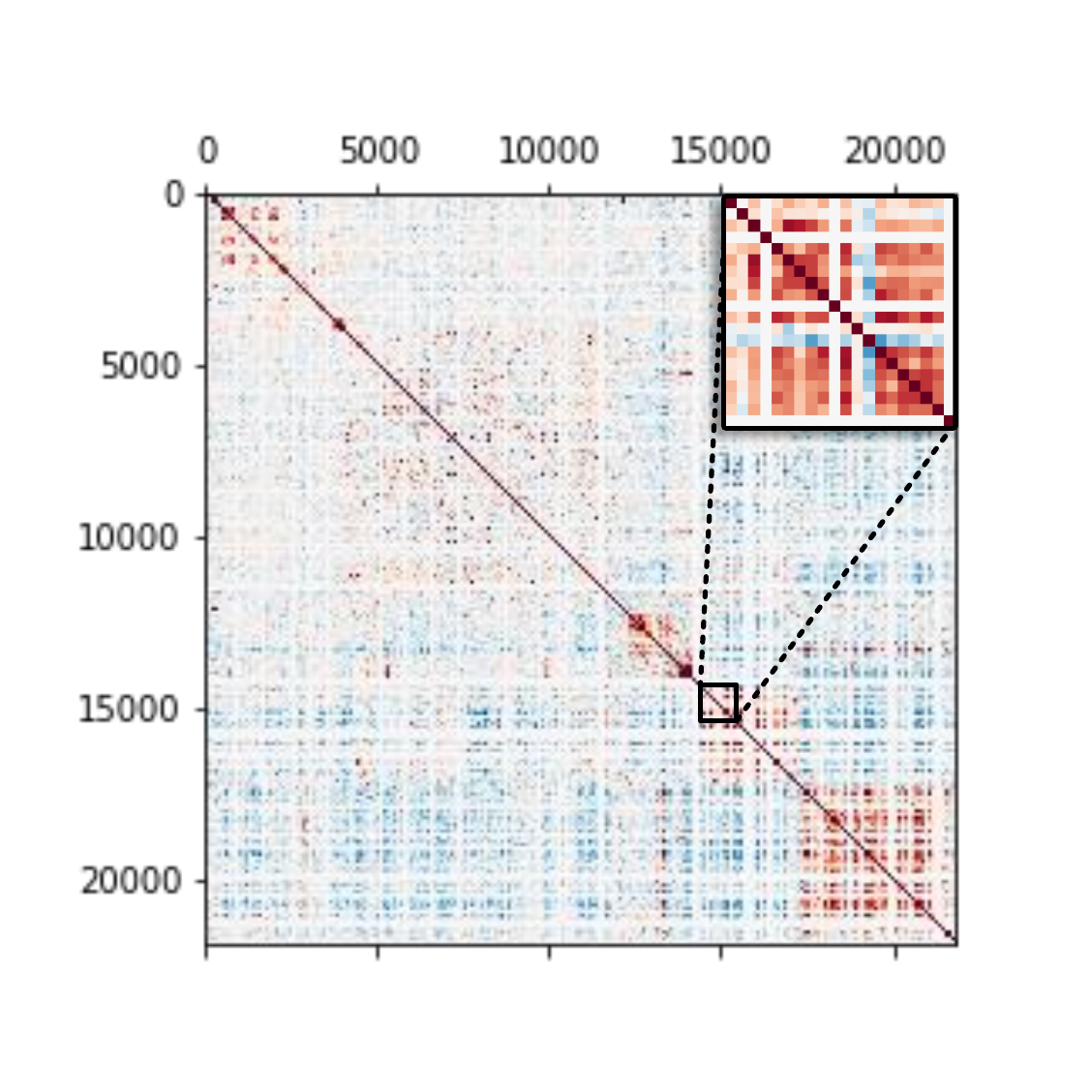}
    }
    \subfloat[TinyViT w/o distill.]{
        \includegraphics[height=2.9cm, trim=20 50 20 50]{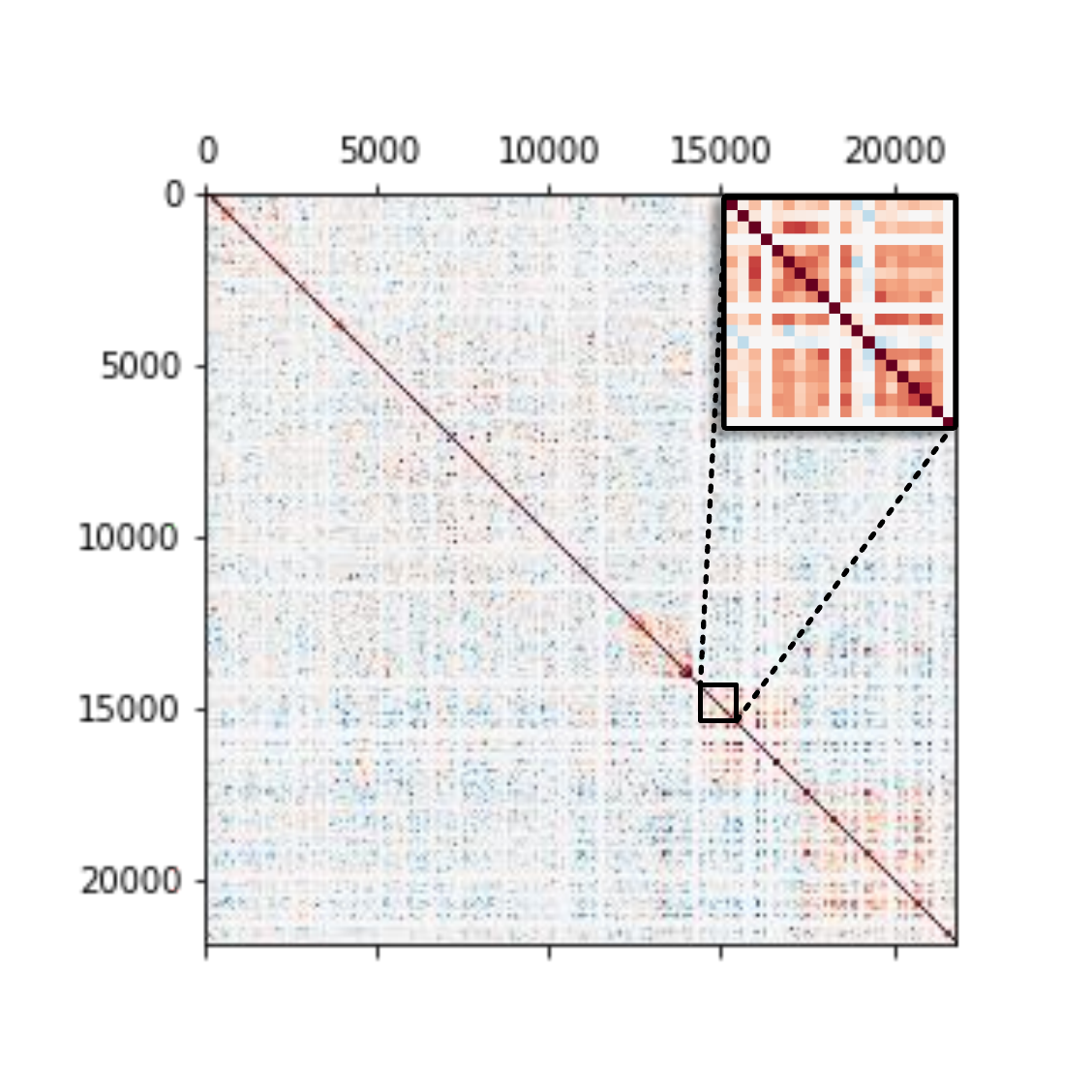}
    }
    \subfloat[TinyViT w/ distill.]{
        \includegraphics[height=2.9cm, trim=20 50 20 50]{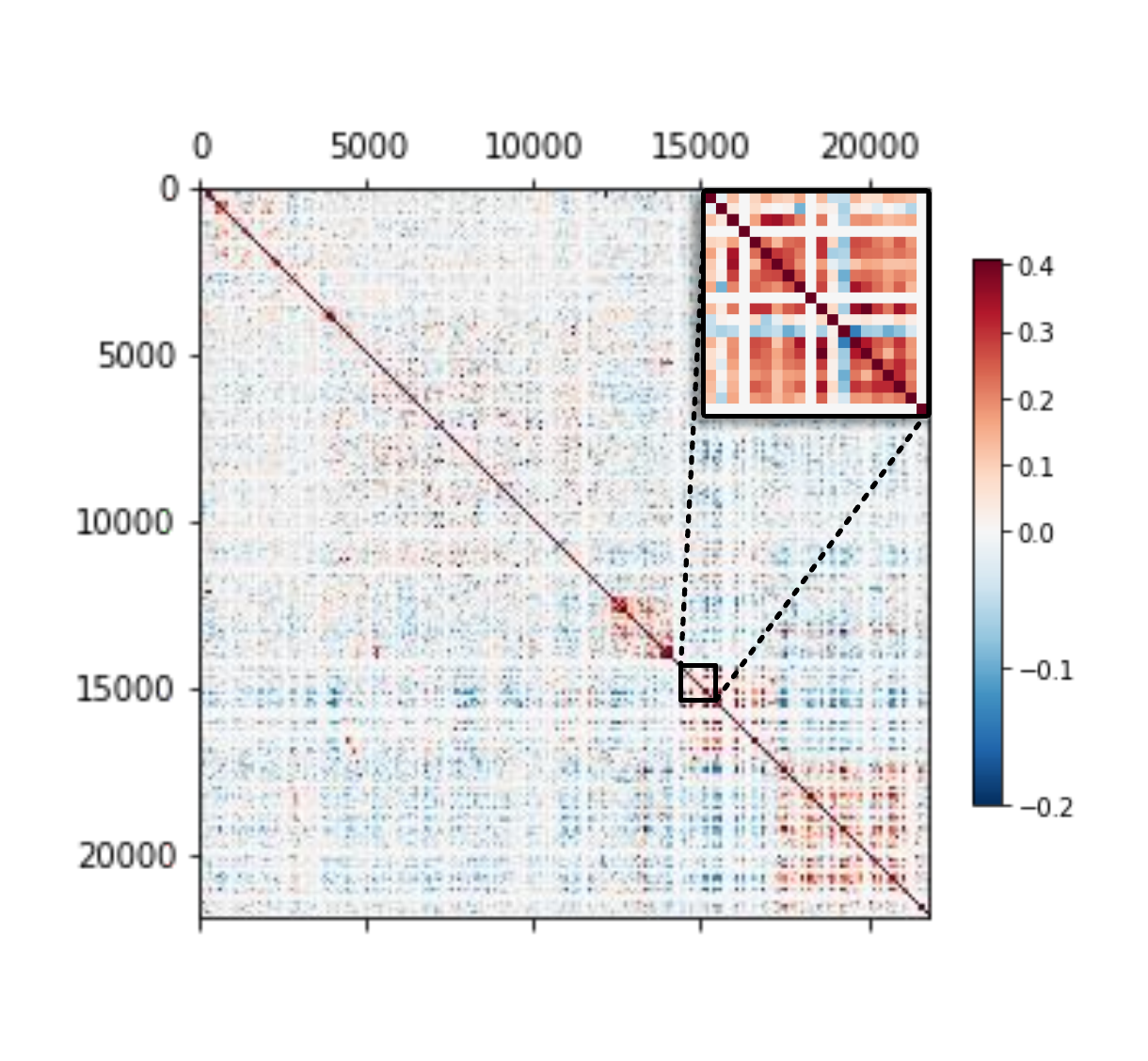}
    }
    \end{minipage}
    \caption {Pearson correlations of output predictions  on ImageNet-21k.}
    \vspace{-4mm}
    \label {fig:vis_map}
\end{figure*}

\textit{Why can distillation improve the performance of small models on large datasets?} The answer is that the student models can directly learn domain knowledge from teachers. Namely, the teacher injects class relationship prior when training the student, while filtering noisy labels (hard samples) for small student models. 

To analyze the class relationships of teacher predictions, 
we select 8 images per class from IN-21k with totally 21,841 classes. These images are then fed into Florence~\cite{florence} to extract prediction logits. Following \cite{understandKD}, we can generate the heatmap of Pearson correlation coefficients between classes on the prediction logits.
In Fig.~\ref{fig:vis_map}(a), simialr or related classes clearly have a high correlations with each other (\textcolor{red}{red}), illustrated by the block diagonal structure. In addition, the teacher model can also capture uncorrelated classes (shown in \textcolor{light_blue}{blue}). This observation verifies that teacher predictions indeed reveal class relationships.

We compare the Pearson correlations on the predictions of TinyViT-21M w/o and w/ distillation, as shown in Fig.~\ref{fig:vis_map}(b) and Fig.~\ref{fig:vis_map}(c) respectively. The block diagonal structure is less obvious without distillation, indicating that the small model is difficult to capture more class relations. However, distillation can guide the student model to imitate the teacher behaviors, leading to better excavating knowledge from large datasets. As shown in Fig.~\ref{fig:vis_map}(c), the Pearson correlations of TinyViT with distillation are closer to the teacher. 

\vspace{-3mm}
\section{Experiments}
\vspace{-2mm}

In this section, we first provide ablation studies on 
 our proposed fast pretraining distillation framework. Next, we compare our TinyViT with other state-of-the-art models. At last, we demonstrate the transferability on downstream tasks.

\vspace{-4mm}
\subsection{\textcolor{black}{Implementation Details}}
\textit{ImageNet-21k pretraining.}
We pretrain TinyViT for 90 epochs on ImageNet-21k~\cite{imagenet} with an AdamW~\cite{adamw} optimizer, a weight decay of 0.01, initial learning rate of 0.002 with a cosine scheduler, 5 epochs warm-up, batch size of 4,096 and gradient clipping with a max norm of 5. The stochastic depth~\cite{stochastic_depth} rate is set to 0 for TinyViT-5/11M and 0.1 for 21M, respectively.
The data augmentation techniques include random resize and crop, horizontal flip, color jittering, random erasing~\cite{random_erase}, RandAugment~\cite{randaug}, Mixup~\cite{mixup} and Cutmix~\cite{cutmix}.

\textit{ImageNet-1k finetuning from the pretrained model.}
We finetune the pretrained models for 30 epochs on ImageNet-1k, using a batch size of 1,024, a cosine learning rate scheduler with 5-epoch warm-up. The initial learning rate is $5\times10^{-4}$ and weight decay is $10^{-8}$. The learning rate of each layer is decayed by the rate 0.8 from the output layer to the input layer. The running statistics of BatchNorm are frozen. We disable Mixup and Cutmix.

\textit{ImageNet-1k finetuning on higher resolution.} When finetuning TinyViT on higher resolution, the windows of each self-attention layer are enlarged as the increasing of input resolution. The attention biases are bilinear-interpolated to adapt the new window size. For example, the window sizes of the four stages are \{7, 7, 14, 7\} on $224^2$ resolution, \{12, 12, 24, 12\} on $384^2$ resolution and \{16, 16, 32, 16\} on $512^2$ resolution.
We finetune the model for 30 epochs, using an accumulated batch size of 1024, a cosine learning rate scheduler with 5-epoch warm up. The initial learning rate is $4\times10^{-5}$ and weight decay is $10^{-8}$. The running statistic of BatchNorm are frozen. Mixup and Cutmix are disabled.

\textit{ImageNet-1k training from scratch.}
We train our models for 300 epochs on ImageNet-1k with an AdamW optimizer, a weight decay of 0.05, initial learning rate of 0.001 with a cosine scheduler, 20 warm-up epochs, batch size of 1,024 and gradient clipping with a max norm of 5.0.
The stochastic depth rate is set to 0.0/0.1/0.2 for TinyViT-5/11M/21M, respectively.

\textit{Knowledge distillation.} We pre-store the top-100 logits of teacher models for IN-21k, including Swin-L~\cite{swin}, BEiT-L~\cite{BEiT}, CLIP-ViT-L/14~\cite{clip,ViT} and Florence~\cite{florence} for all 90 epochs. Note that CLIP-ViT-L/14 and Florence are finetuned on IN-21k for 30 epochs to serve as teachers. Then, we distill the student models using the stored teacher logits with the same hyper-parameters as the distillation involving the teacher model. The distillation temperature is set to 1.0. We disable Mixup~\cite{mixup} and Cutmix~\cite{cutmix} for pretraining distillation on TinyViT.
All models are implemented using PyTorch~\cite{pytorch} with timm library~\cite{timm}.

\vspace{-3mm}
\subsection{Ablation Study}
\label{sec:ablation}

\begin{table}[t]
\begin{center}
\caption{Ablation study on different pretraining strategies for Swin~\cite{swin} and DeiT~\cite{deit}. The performance on IN-1k is reported.
}
\label{table:swint}
\setlength{\tabcolsep}{15pt}{
\scalebox{0.9}{
\begin{tabular}{c|c|c|cc}
\Xhline{2\arrayrulewidth}
\multirow{2}{*}{Model} & \#Params & Train on & \multicolumn{2}{c}{Pretrain on IN-21k} \\
~ & (M) &IN-1k & w/o distill. & w/ distill. \\
\Xhline{2\arrayrulewidth}
DeiT-Ti~\cite{deit} & 5 & 72.2 &  73.0{\textcolor{YellowGreen}{\scriptsize{(+0.8)}}} & 74.4{\textcolor{YellowGreen}{\scriptsize{(+2.2)}}} \\
DeiT-S~\cite{deit} & 22 & 79.9 & 80.5{\textcolor{YellowGreen}{\scriptsize{(+0.6)}}} & 82.0{\textcolor{YellowGreen}{\scriptsize{(+2.1)}}} \\
Swin-T~\cite{swin} & 28 & 81.2 & 81.9{\textcolor{YellowGreen}{\scriptsize{(+0.7)}}} & 83.4{\textcolor{YellowGreen}{\scriptsize{(+2.2)}}} \\
\Xhline{2\arrayrulewidth}
\end{tabular}}}
\end{center}
\vspace{-5mm}
\end{table}

\textit{Impact of pretraining distillation on existing small ViTs.} We study the effectiveness of our proposed fast pretraining distillation framework on two popular vision transformers: DeiT~\cite{deit} and Swin~\cite{swin}. As shown in Tab.~\ref{table:swint}, comparing to training from scratch on IN-1k, pretraining without distillation on IN-21k can only bring limited gains, \emph{i.e.} 0.8\%/0.6\%/0.7\% for DeiT-Ti/DeiT-S/Swin-T, respectively. However, our proposed fast pretraining distillation framework increases the accuracy by 2.2\%/2.1\%/2.2\% respectively.
\REV{It indicates that pretraining distillation allows small models to benefit more from large-scale datasets. }

\begin{figure*}[t]
  \vspace{-2mm}
  \centering
  
  \begin{subfigure}{0.49\linewidth}
    \includegraphics[width=1.0\linewidth, trim=0 50 0 0]{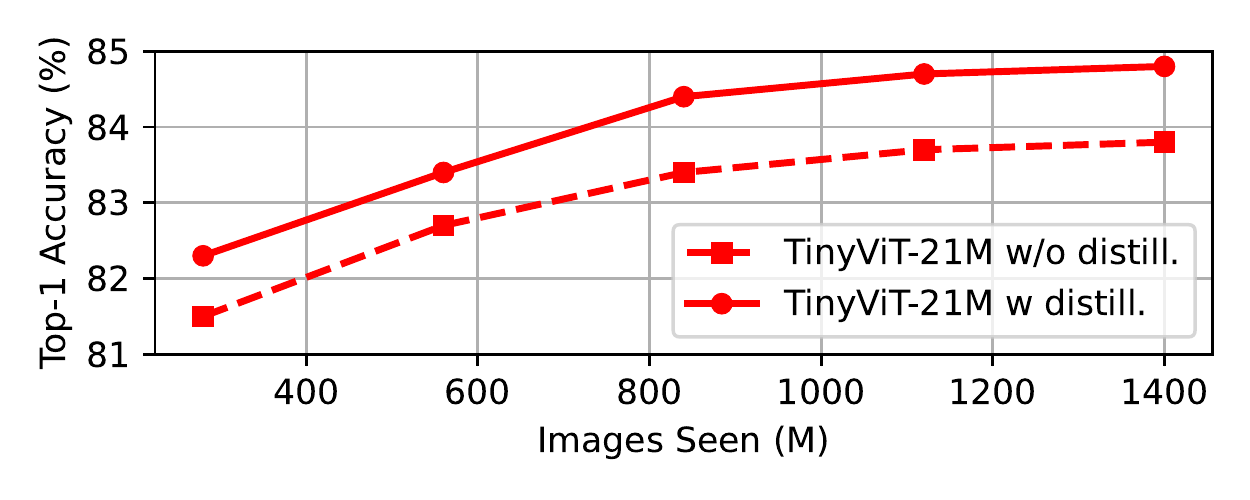}
    \label{fig:data_scale_1}
  \end{subfigure}
  \hfill
  \begin{subfigure}{0.49\linewidth}
    \includegraphics[width=1.0\linewidth, trim=0 50 0 0]{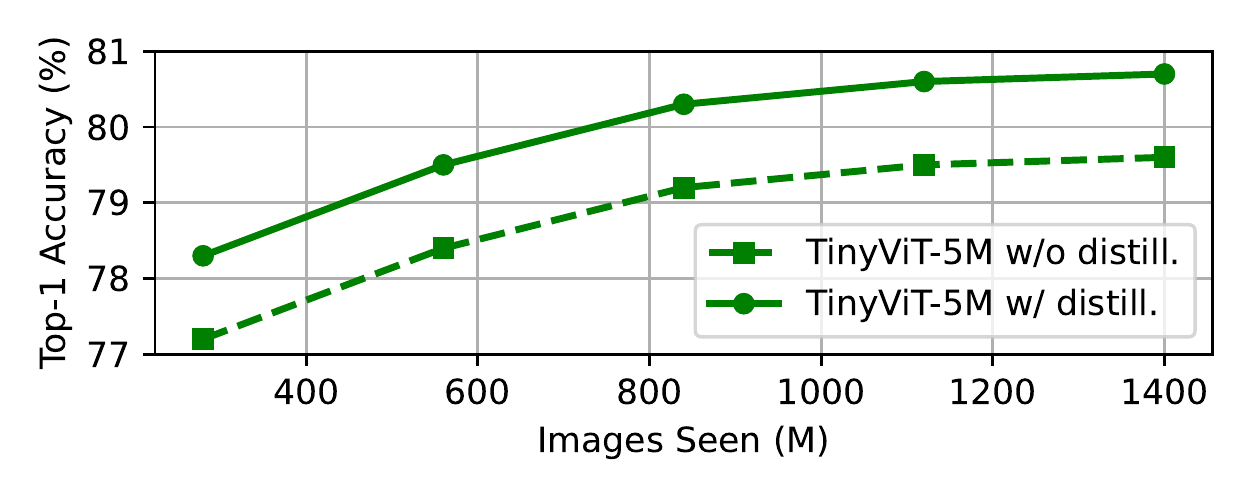}
    \label{fig:data_scale_2}
  \end{subfigure}
  
  \caption{Comparison on pretrained TinyViT-21M/5M over training data size.}
  \label{fig:data_scale}
  \vspace{-4mm}
\end{figure*}

\begin{figure*}[t]
  \centering
  
  \begin{subfigure}{0.49\linewidth}
    \includegraphics[width=1.0\linewidth, trim=0 50 0 30]{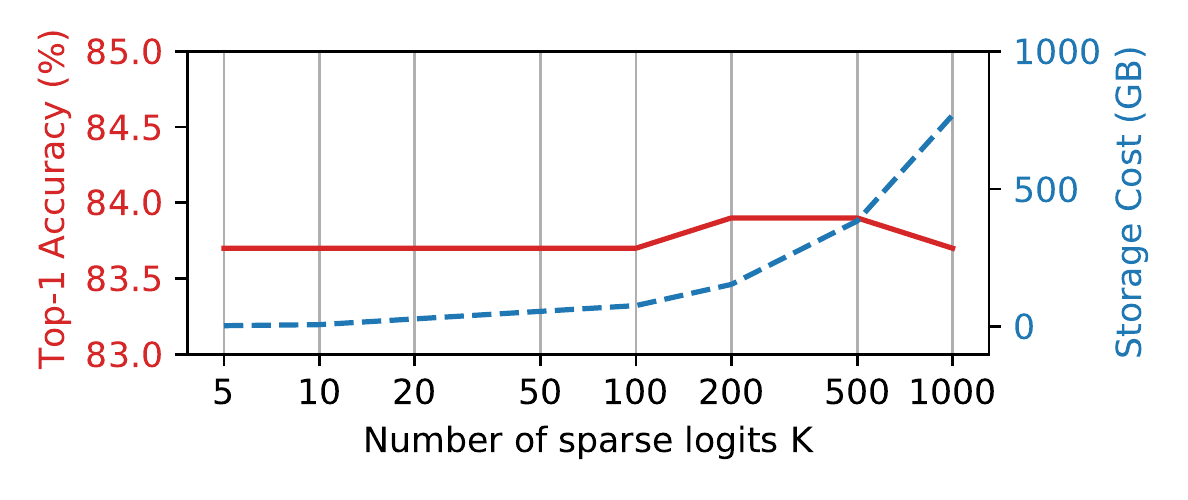}
    \label{fig:topk-1k}
  \end{subfigure}
  \hfill
  \begin{subfigure}{0.49\linewidth}
    \includegraphics[width=1.0\linewidth, trim=0 50 0 30]{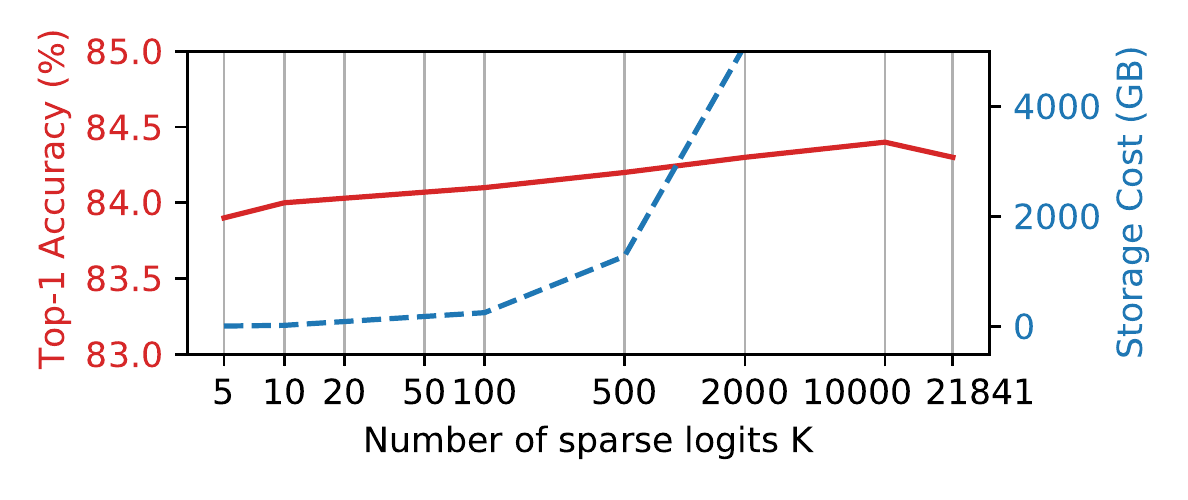}
    \label{fig:topk-21k}
  \end{subfigure}
  
  \caption{The accuracy on IN-1k and storage cost of TinyViT-21M along different saved logits $K$. \textbf{Left: } distill TinyViT-21M on IN-1k
  \textbf{Right: } distill TinyViT-21M on IN-21k then finetune it on IN-1k.
  }
  \label{fig:topk}
  \vspace{-4mm}
\end{figure*}

\textit{Impact of pretraining data scale.}
We investigate the representation quality of TinyViT-5M/21M with respect to the total number of images ``seen" (batch size times number of steps) during pretraining on IN-21k, following the strategies in \cite{scalevit}. We use CLIP-ViT-L/14~\cite{clip,ViT} as the teacher. The results on IN-1k after finetuning are shown in Fig.~\ref{fig:data_scale}. We have the following observations. 1) For both models, pretraining distillation can consistently brings performance gains over different data size. 
2) All models tends to saturate as the number of epochs increase, which may be bottlenecked by the model capacities. 

\textit{Impact of the number of saved logits.}
The effects of sparse logits on distilling TinyViT-21M by using Swin-L~\cite{swin} as the teacher model are shown in Fig.~\ref{fig:topk}.
On both IN-1k and IN-21k, \REV{we observe that the accuracy increases as the number of sparse logits $K$ grows until saturation, meanwhile the storage cost grows linearly.}

This observation is aligned with existing work on knowledge distillation~\cite{understandKD,fkd}, where teacher logtis capture class relationships but also contain noise.
This makes it possible to sparsify teacher logits such that the class relationships are reserved while reducing noise.
Moreover, memory consumption also impose constraints on the choice of $K$.
\REV{To obtain comparable accuracy under limited storage space, we select the slightly larger $K$, where $K$=10~(1.0\% logits) on IN-1k for 300 epochs and $K$=100~(0.46\% logits) on IN-21k for 90 epochs using 16~GB/481~GB storage cost, respectively.}

\textit{Impact of teacher models.}
We evaluate the impact of teacher models for pretraining distillation.
As shown in Tab.~\ref{table:headings}, a better teacher can yield better student models (\#1 \emph{vs.} \#2 \emph{vs.} \#3 and \#4).
TinyViT-21M distilled by Florence on IN-21k is 1.0\%/0.6\%/1.0\% higher in top-1 accuracy on three benchmark datasets than trained from scratch on IN-21k (\#0 \emph{vs.} \#4). 
However, better teacher models are often large in model size, resulting in high GPU memory consumption and long training time, e.g., Florence (\#4) with 682M parameters occupies 11GB GPU memory and leads to 2.4 times longer training time.

Note that our fast pretraining distillation framework simply loads the teacher logits from a hard disk during training. Therefore, it does not require additional GPU memory and has the same training time as \#0.  Moreover, the framework is compatible with all types of teacher models. Therefore, the performance of TinyViT can be further improved by introducing more powerful teachers.

\begin{table}[t]
\vspace{-3mm}
\begin{center}
\caption{ Ablation study on different teacher models for pretraining distillation.
Teacher performance are listed in the brackets: (the number of parameters, linear probe performance on IN-1k). We report the training time cost and memory consumption of teacher models on NVIDIA V100 GPUs without using our proposed fast pretraining distillation.
}
\label{table:headings}
\setlength{\tabcolsep}{4pt}{
\scalebox{0.75}{
\begin{tabular}{c|c|ccc|cc}
\Xhline{2\arrayrulewidth}
\multirow{2}{*}{\#} & IN-21k &IN-1k & IN-Real & IN-V2 & Training Time & Memory\\
& Pretrained Teacher & Top-1(\%)&Top-1(\%) &Top-1(\%) & (GPU Hours) & (GB)\\

\Xhline{2\arrayrulewidth}

0 & w/o distill. & 83.8 & 88.4 & 73.8 & 3,360 & 0 \\
\hline

1 & BEiT-L (326M, 84.1)~\cite{BEiT} & 84.1 & 88.4 & 73.8 & 6,415~\textcolor{light_blue}{($1.9\times$)} & 3.9 \\
2 & Swin-L (229M, 84.4)~\cite{swin} & 84.2 & 88.6 & 73.9 & 5,804~\textcolor{light_blue}{($1.7\times$)} & 6.8 \\

3 & CLIP-ViT-L/14 (321M, 85.2)~\cite{clip} & 84.8 & 88.9 & 75.1 & 7,087~\textcolor{light_blue}{($2.1\times$)} & 2.7 \\
4 & Florence (682M, 86.2)~\cite{florence} & 84.8 & 89.0 & 74.8 & 7,942~\textcolor{light_blue}{($2.4\times$)} & 10.7 \\
\Xhline{2\arrayrulewidth}
\end{tabular}}}
\end{center}
\vspace{-2mm}
\end{table}

\begin{figure*}[t]
  \centering
  \begin{subfigure}{0.49\linewidth}
    \includegraphics[height=0.478\linewidth,trim=0 50 0 30]{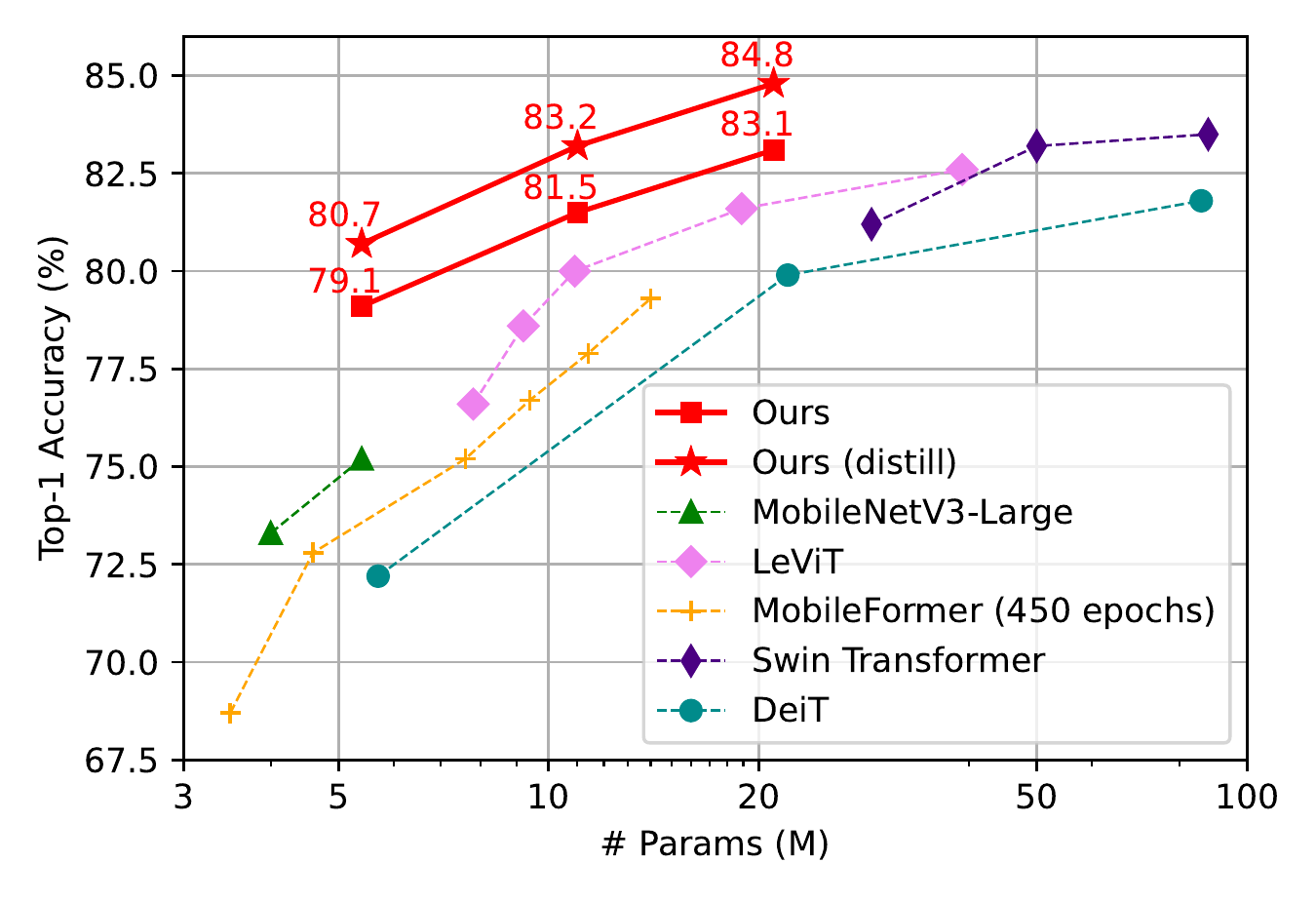}
    \label{fig:short-a}
  \end{subfigure}
  \hfill
  \begin{subfigure}{0.49\linewidth}
    \includegraphics[height=0.485\linewidth,trim=10 52 0 28.5]{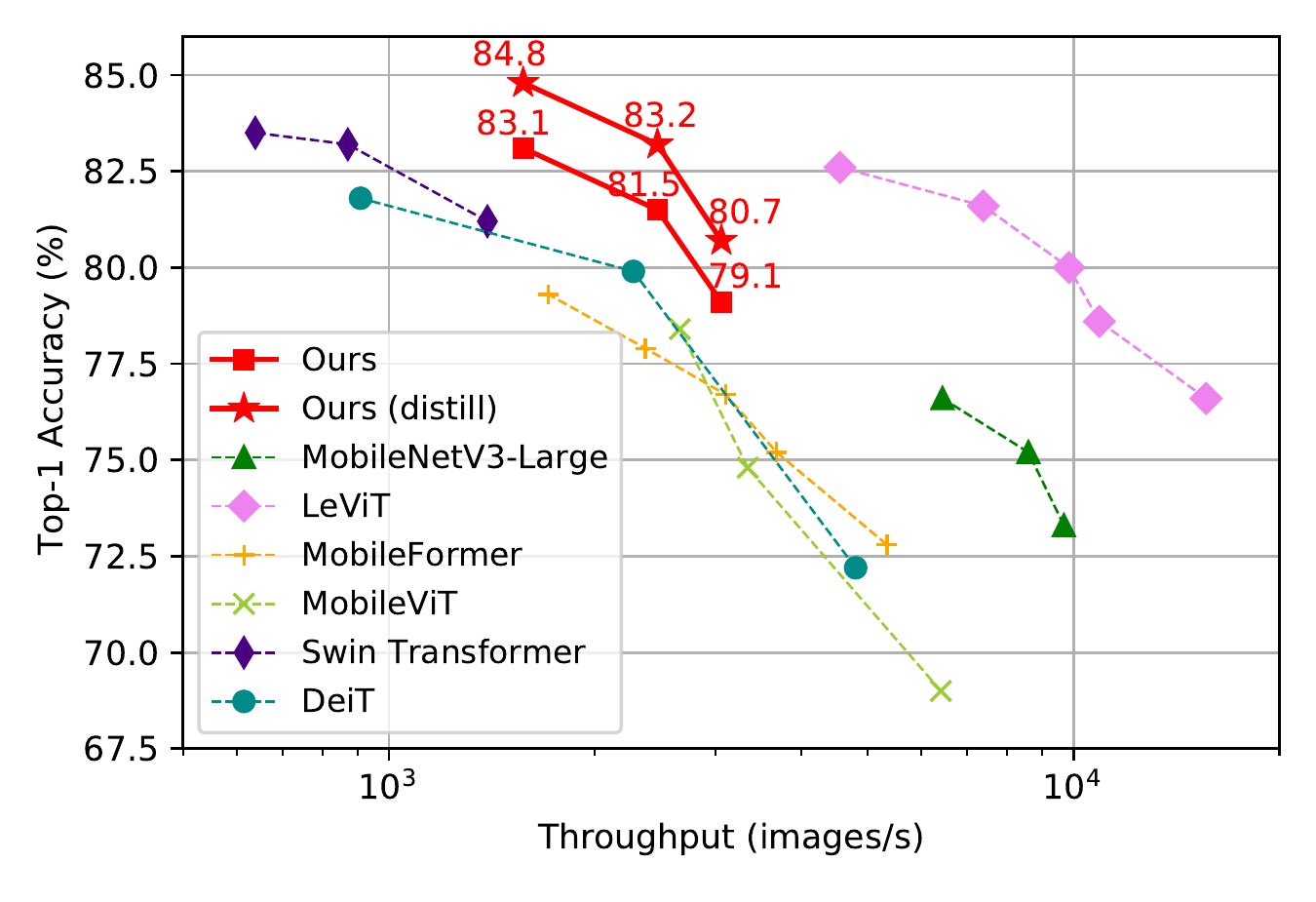}
    \label{fig:short-b}
  \end{subfigure}
  \caption{Comparison
  with state-of-the-art methods
  on IN-1k.}
  \label{fig:acc}
  \vspace{-4mm}
\end{figure*}

\begin{table*}[t]
\vspace{-4mm}
\centering

\caption {
TinyViT performance on IN-1k~\cite{imagenet} with comparisons to state-of-the-art models. MACs~(multiply–accumulate operations) and Throughput are measured using the GitHub repository of~\cite{fvcore,levit} and a V100 GPU. \REV{\alambic: pretrain on IN-21k with the proposed fast distillation; $\uparrow$: finetune with higher resolution.}
}
    
\setlength{\tabcolsep}{4pt}{
\scalebox{0.75}{

\begin{tabular}{c|l||cc||ccc||cc}

    \toprule[1.5pt]
    & \multirow{2}{*}{Model} & Top-1 & Top-5 & \#Params & MACs & Throughput & \multirow{2}{*}{Input} & \multirow{2}{*}{Arch.} \\
    &  & (\%) & (\%) & (M) & (G) & (images/s) &  &  \\
    \midrule[1.5pt]
    \multirow{7}{*}{\rotatebox[origin=l]{90}{5-10M \#Params}} 
    
    & MoblieViT-S~\cite{mobilevit} &78.4&-& 6 & 1.8 & 2,661 & 256 & Hybrid \\
    & ViTAS-DeiT-A~\cite{vitas} & 75.5 & 92.4 & 6 & 1.3 & 3,504 & 224 & Trans \\
    
    & GLiT-Tiny~\cite{glit} & 76.3 & - & 7 & 1.5 & 3,262 & 224 & Trans \\
    
    & Mobile-Former-214M~\cite{mobileformer} & 76.7 & - & 9 & 0.2 & 3,105 & 224 & Hybrid \\
    & CrossViT-9~\cite{CrossViT} & 77.1 & - & 9 & 2.0 & 2,659 & 224 & Trans \\

    & \textbf{TinyViT-5M}~(\textbf{ours})& \textbf{79.1}& 94.8 & 5.4 & 1.3 & 3,060 &224 & Hybrid\\
    & \textbf{TinyViT-5M}\alambic~(\textbf{ours}) &\textbf{80.7} & 95.6 & 5.4 & 1.3 & 3,060 &224 & Hybrid\\
    
    \midrule[1.5pt]
    
    \multirow{7}{*}{\rotatebox[origin=l]{90}{11-20M}}

    & ResNet-18~\cite{resnet} & 70.3 & 86.7 & 12 & 1.8 & 8,714 & 224 & CNN \\
    & PVT-Tiny~\cite{pvt} & 75.1 & - & 13 & 1.9 & 2,791 & 224 & Trans \\
    & ResT-Small~\cite{rest} & 79.6 & 94.9 & 14 & 2.1 & 2,037 & 224 & Trans \\
    & LeViT-256~\cite{levit} & 81.6 & - & 19& 1.1 & 7,386 & 224 & Hybrid \\
     & CoaT-Lite Small~\cite{coat} & 81.9 & 95.6 & 20 & 4.0 & 1,138 & 224 & Trans \\
    
    & \textbf{TinyViT-11M}~(\textbf{ours})& \textbf{81.5} & 95.8 & 11 & 2.0 & 2,468 &224 & Hybrid\\
    & \textbf{TinyViT-11M}\alambic~(\textbf{ours})&\textbf{83.2} & 96.5 & 11 & 2.0 & 2,468 &224 & Hybrid\\
    
    \midrule[1.5pt]
    \multirow{7}{*}{\rotatebox[origin=l]{90}{$>$20M}} 
    
    & DeiT-S~\cite{deit} & 79.9& 95.0 &22 & 4.6 & 2,276 &224 & Trans\\
    & T2T-ViT-14~\cite{T2TViT} & 81.5 & 95.7 & 21 & 4.8& 1,557 &224 & Trans\\ 
    & AutoFormer-S~\cite{autoformer} &81.7&95.7& 23 & 5.1& 1,341 & 224 & Trans \\

    & Swin-T~\cite{swin} &81.2& 95.5 & 28 &4.5 & 1,393 & 224 & Trans \\
    & CrossViT-15~\cite{CrossViT} & 82.3 & - & 28 & 6.1 & 1,306 & 224 & Trans \\
    & EffNet-B5~\cite{efficientnet} &83.6&96.7& 30 & 9.9 & 330 &  456& CNN \\

    & \textbf{TinyViT-21M}~(\textbf{ours})& \textbf{83.1} & 96.5 & 21 & 4.3 & 1,571 &224 & Hybrid\\
    & \textbf{TinyViT-21M}\alambic~(\textbf{ours})& \textbf{84.8} & 97.3  & 21 & 4.3 & 1,571 &224 & Hybrid\\
    & \textbf{TinyViT-21M\alambic$\uparrow$384}~(\textbf{ours})& \textbf{86.2} & 97.8 & 21 & 13.8 & 394 & 384 & Hybrid\\
    & \textbf{TinyViT-21M\alambic$\uparrow$512}~(\textbf{ours})& \textbf{86.5} & 97.9 & 21 & 27.0 & 167 & 512 & Hybrid\\
    
    \bottomrule[1.5pt]
    \end{tabular}
    
}}
    
    \label {tab:sota_cls}
    \vspace{-4mm}
\end{table*}

\vspace{-4mm}
\subsection{Results on ImageNet}
In this section, we compare our scaled TinyViT models with state-of-the-art methods on IN-1k~\cite{imagenet}. 
The performance is reported in Fig.~\ref{fig:acc} and Tab.~\ref{tab:sota_cls}.
The models with \alambic indicates pretraining on IN-21k with the proposed fast distillation framework
using CLIP-ViT-L/14~\cite{clip,ViT} as the teacher. It shows that, without distillation, our TinyViT models achieve comparable performance to current prevailing methods, such as Swin transformer~\cite{swin} and LeViT~\cite{levit}, with similar parameters. This indicates the effectiveness of the proposed new architectures and the model scaling techniques. Moreover, with the fast pretraining distillation, the performance of TinyViT can be largely improved, outperforming the state-of-the-art CNN, transformer and hybrid models. In particular, using only 21M parameters, TinyViT trained from scratch on IN-1k gets 1.9\%/3.2\% higher top-1 accuracy than Swin-T~\cite{swin} and DeiT-S~\cite{deit} respectively, while after pretraining with distillation on IN-21k, the improvements arise to 3.6\% and 4.9\%. With higher resolution, TinyViT-21M reaches a top-1 accuracy of 86.5\%, establishing new state-of-the-art performance on IN-1k for small models. Besides, TinyViT surpasses automatically searched models, such as AutoFormer~\cite{autoformer} and GLiT~\cite{glit}.

\begin{table}[t]
\vspace{-2mm}
      \centering
      \caption{Performance of TinyViT-21M
      w/ and w/o pretraining 
      for linear probe and few-shot image classification. 
      }
      \setlength{\tabcolsep}{4pt}{
      \scalebox{0.7}{
      \begin{tabular}{c|c|ccccc|cccc|cccc|cccc}
      \Xhline{2\arrayrulewidth}
         
      \multirow{2}{*}{\makecell{\\~\\~\\ \#\\}} & \multirow{2}{*}{\makecell{\\~\\~\\Training\\dataset}} & \multicolumn{5}{c|}{Linear probe} & \multicolumn{4}{c|}{5-shot} & \multicolumn{4}{c|}{20-shot} & \multicolumn{4}{c}{50-shot} \\
        \Xhline{2\arrayrulewidth}
       &  & \rotatebox[origin=l]{90}{CIFAR-10} & \rotatebox[origin=l]{90}{CIFAR-100  } & \rotatebox[origin=l]{90}{Flowers} & \rotatebox[origin=l]{90}{Cars} & \rotatebox[origin=l]{90}{Pets} & \rotatebox[origin=l]{90}{ISIC} & \rotatebox[origin=l]{90}{EuroSAT} & \rotatebox[origin=l]{90}{CropD} & \rotatebox[origin=l]{90}{ChestX} & \rotatebox[origin=l]{90}{ISIC} & \rotatebox[origin=l]{90}{EuroSAT} & \rotatebox[origin=l]{90}{CropD} & \rotatebox[origin=l]{90}{ChestX} & \rotatebox[origin=l]{90}{ISIC} & \rotatebox[origin=l]{90}{EuroSAT} & \rotatebox[origin=l]{90}{CropD} & \rotatebox[origin=l]{90}{ChestX} \\
        
    \Xhline{2\arrayrulewidth}
        
      0 & IN-1k & 91.7 & 75.2 & 80.9 & 56.3 & 86.5
        & 42.9 & 82.4 & 92.2 & 24.8
        & 56.7 & 91.0 & 97.4 & 29.0
        & 63.7 & 94.2 & 98.6 & 31.8 \\

       1 & IN-1k\alambic & 91.7 & 74.5 & 82.4 & 61.7 & 85.5
        & 43.0 & 83.0 & 94.2 & 24.4 
        & 58.5 & 91.8 & 97.9 & 28.6 
        & 66.2 & 94.3 & 98.9 & 31.8 \\

       2 & IN-21k & 96.3 & 84.7 & 99.7 & 67.7 & 92.6
        & 52.5 & 87.4 & 97.4 & 24.6
        & 66.5 & 93.7 & 99.1 & 29.4
        & 73.4 & 95.5 & 99.5 & 33.4 \\

       3 & IN-21k\alambic & 96.9 & 86.6 & 99.7 & 75.1 & 93.8
        & 53.5 & 88.1 & 98.0 & 24.7
        & 67.3 & 93.9 & 99.3 & 29.5
        & 74.2 & 96.0 & 99.5 & 33.2 \\
        
        \Xhline{2\arrayrulewidth}
      \end{tabular}
      }}
      \label{tab:downstream_task}
      \vspace{-4mm}
\end{table}

\vspace{-3mm}
\subsection{Transfer Learning Results}

\quad\quad\textbf{Linear Probe.} For linear probe, we follow the same setting as in MOCO v3 \cite{mocov3}, \emph{i.e.}, replacing the head of TinyViT models with a linear layer, while only finetuning the linear layer on downstream datasets and frozing other weights. We consider five classification benchmarks: CIFAR-10~\cite{cifar}, CIFAR-100~\cite{cifar}, Flowers~\cite{flowers}, Cars~\cite{stanford_cars} and Pets~\cite{pet}. The results are reported in Tab. \ref{tab:downstream_task}. 

We compare the performance of TinyViT-21M with 4 different training settings. It is clear that distillation can improve the linear probe performance of TinyViT (\#0 \emph{vs.} \#1, \#2 \emph{vs.} \#3). Besides, when trained on larger datasets (\emph{i.e.}, IN-21k), TinyViT gets more than 10\% gains over CIFAR-100, Flowers and Cars (\#0,\#1 \emph{vs.} \#2, \#3), indicating better representability. Thus, pretraining with distillation on large-scale datasets achieves the best representability (\#3).

\textbf{Few-shot Learning.}
We also evaluate the transferability of TinyViT with different training settings on few-shot learning benchmark~\cite{fewshot}. 
The benchmark datasets include: \REV{CropDisease \cite{cropdisease} (plant leaf images, 38 disease stages over 14 plant species), EuroSAT~\cite{eurosat} (RGB satellite images, 10 categories), ISIC 2018~\cite{isic} (dermoscopic images of skin lesions, 7 disease states) and ChestX~\cite{chestx} (Chest X-rays, 16 conditions). The learning and inference settings are the same as in \cite{fewshot}. The evaluation protocol involves 5-way classification across 5-, 20- and 50-shot. The classes and shots are randomly sampled for each episode, for 600 episodes per way and shot. Average accuracy over all episodes is reported. We add a single linear layer in replace of the original classification layer in TinyViT-21M. 
As shown in Tab.~\ref{tab:downstream_task},
we obtain same observations as the linear probe results, except of ChestX, where gray-scale medical images are the least similar to natural images, as well as few in the training dataset for the teacher models and the student models.}
In combination of these results, we can conclude that pretraining distillation is significant in improving the representability of small models, and thus our proposed fast pretraining distillation framework is effective.

\textbf{Object Detection.} We also investigate the transfer ability of our TinyViT on object detection task~\cite{coco}. We use Cascade R-CNN~\cite{cascade_rcnn} with Swin-T~\cite{swin} as our baseline. We follow the same training settings used in Swin transformer~\cite{swin}. The results on COCO 2017 validation set are reported in Tab.~\ref{tab:det_comp}. Under the same training recipe, our TinyViT architecture achieves better performance than Swin-T, getting 1.5\% AP improvements. Furthermore, after applying pretraining distillation, TinyViT gets another 0.6\% AP improvements, being 2.1\%  higher than Swin-T. This clearly demonstrates our fast pretraining distillation framework is effective and capable of improving the transfer ability of small models.

 \begin{table*}[t]
    \caption {Comparison on COCO~\cite{coco} object 
    detection using Cascade Mask R-CNN~\cite{cascade_rcnn,mask_rcnn} \REV{for 12 epochs. 
    We report the number of parameters of the backbone.}
    }
    \centering
      \setlength{\tabcolsep}{6pt}{
      \scalebox{0.85}{
      
    \begin{tabular}{c|cc|c|cccccc}
       \Xhline{2\arrayrulewidth}
        \# & Backbone & \#Params & IN-1k& $AP$ & $AP_{50}$ & $AP_{75}$ & $AP_S$ & $AP_M$ & $AP_L$ \\
       \Xhline{2\arrayrulewidth}
        0  & Swin-T~\cite{swin} & 28M & 81.2 &  48.1 & 67.1 & 52.1 & 31.1 & 51.2 & 63.5 \\
       \hline
        1 & TinyViT-21M & 21M  & 83.1 & 49.6 {\textcolor{YellowGreen}{\scriptsize{(+1.5)}}} & 68.5 & 54.2 & 32.3 & 53.2 & 64.8\\

        2 & TinyViT-21M\alambic & 21M & 84.8 & 50.2 {\textcolor{YellowGreen}{\scriptsize{(+2.1)}}} & 69.4 & 54.4 & 32.9 & 53.9 & 65.2 \\

       \Xhline{2\arrayrulewidth}
    \end{tabular}}}
    \label {tab:det_comp}
    \vspace{-4mm}
\end{table*}

\vspace{-3mm}
\section{Conclusions}
We have proposed a new family of tiny and efficient vision transformers pretrained on large-scale datasets with our proposed fast distillation framework, named TinyViT.
Extensive experiments demonstrate the efficacy of TinyViT on ImageNet-1k, and its superior transferability on various downstream benchmarks. In future work, we will consider using more data to further unlock the representability of small models with the assistance of more powerful teacher models. Designing a more effective scaling down method to generate small models with better computation/accuracy is another interesting research direction.

\clearpage

\bibliographystyle{splncs04}
\bibliography{egbib}

\title{TinyViT: Fast Pretraining Distillation for \\  Small Vision Transformers \\ ------ Supplementary Material ------ }


\titlerunning{TinyViT}
\author{ }
\authorrunning{Kan Wu et al.}
\institute{ }

\maketitle

This supplementary material presents the details of Section 3.2. Besides, two extra experiments show how fast the proposed distillation method is and the result of distillation with ground-truth.

\begin{itemize}[noitemsep,topsep=0pt,parsep=0pt,partopsep=0pt]

\item \textbf{Model Architectures.} We elaborate the model architectures of TinyViT of Section 3.2.
\item \textbf{How fast the distillation method is?} We compare the training cost between our proposed fast pretraining distillation and the conventional method, to show the effectiveness of the proposed method.
\item \textbf{Distillation with ground-truth.} We show why to use soft labels only to distill student models.

\end{itemize}

\appendix
\section{Model Architectures}
\label{sec:model_arch}

Our proposed TinyViT architecture is shown in Tab.~\ref{tab:arch}. It is a hierarchical structure with 4 stages, for the convenience of dense prediction downstream tasks like Swin~\cite{swin} and LeViT~\cite{levit}. The attention biases~\cite{levit} and a $3\times3$ depthwise convolution between attention and MLP are introduced to capture local information~\cite{irpe,chu2021conditional}.
The factors \{\TdimALL, \dALL, \wsALL, \IBratio, \mlp, \Tdimh\} can be contracted to form tiny model families.
We start with a 21M model and generate a set of candidate models around the basic model by adjusting the contraction factors. Then we select models that satisfy both constraints on the number of parameters and throughput, and evaluate them on 99\% train and 1\% val data sampled from ImageNet-1k train set. The models with the best validation accuracy will be utilized for further reduction in the next step until the target is achieved. In TinyViT model family, all models share the same factors: \{\IBdepth, \dtwo, \dthree, \dfour\} = \{2, 2, 6, 2\}, \{\wstwo, \wsthree, \wsfour\} = \{7, 14, 7\} and \{\IBratio, \mlp, \Tdimh\} = \{4, 4, 32\}. For the embeded dimensions \{\Tdima, \Tdimb, \Tdimc, \Tdimd\}, TinyViT-21M: \{96, 192, 384, 576\}, TinyViT-11M: \{64, 128, 256, 448\} and TinyViT-5M: \{64, 128, 160, 320\}.

Besides, we provide some interesting observations about model contraction. It may help both the manual design and the search space design for efficient small vision transformers.

1) For small vision transformers, it improves the accuracy when replacing the transformer block in the first stage with MBConv~\cite{mobilenetv3} blocks. We conjecture that early convolution introduces inductive bias like locality~\cite{early_conv,levit}. It provides more prior knowledge to help small models converge well.

2) It reduces the number of parameters significantly when decreasing the embeded dimension \TdimALL, so it is the first step to scale the model down. When the model becomes narrower, its depth (especially in the depth of the third stage~\dthree) is increased to satisfy the constraint of the number of parameters.

3) The MLP expansion ratio \mlp~4 is better than 3 in our models. 

4) Window sizes \wsALL~do not affect the model size, but larger windows imrpove the accuracy with more computational cost. Especially for Stage 3, $14\times14$ window size improves the accuracy with little extra computational cost.

\begin{table*}[t]
\vspace{-5mm}
\caption{An elastic base architecture of TinyViT.}
\centering
\setlength{\tabcolsep}{3pt}{
    \scalebox{0.8}{
        \begin{tabular}{c|c|c|c}
        \Xhline{2\arrayrulewidth}
            & Block & Configuration & Output \\
            \Xhline{2\arrayrulewidth}
            Patch Embed &
            Stacked Conv &
            $
                \begin{bmatrix}
                    \text{kernel size }~3\times3, \\
                    \text{stride 2, padding 1}
                \end{bmatrix}
                \times 2
            $
            & $56\times56$ \\
            \hline
            Stage 1 & 
           \text{MBConv}~\cite{mobilenetv3}
             & 
            $
            \begin{bmatrix} 
            \text{embed dim }\TdimaRaw, \\
            \text{expansion ratio }\IBratioRaw
            \end{bmatrix}
            \times\IBdepthRaw$
            
            & $56\times56$ \\
            \hline
            Downsampling &
            \text{MBConv}~\cite{mobilenetv3}
            &
            $
            \begin{bmatrix}
                \text{embed dim } \TdimaRaw, \text{stride }2, \\
                \text{hidden/output dim } \TdimbRaw
            \end{bmatrix}\times 1
            $
            & $28\times28$ \\
            \hline
            Stage 2 & 
            Transformer~\cite{vaswani2017attention} &
            $
            \begin{bmatrix}
               & \text{embed dim }\TdimbRaw,~\text{head }\TdimbRaw/\TdimhRaw, &\\
               & \text{window size }\wstwoRaw\times\wstwoRaw, &\\
               & \text{mlp ratio }\mlpRaw &
            \end{bmatrix}
            \times \dtwoRaw$
            & $28\times28$ \\
            \hline
            Downsampling &
            \text{MBConv}~\cite{mobilenetv3}
            &
            $
            \begin{bmatrix}
                \text{embed dim } \TdimbRaw, \text{stride }2, \\
                \text{hidden/output dim } \TdimcRaw
            \end{bmatrix}\times 1
            $
            & $14\times14$ \\
            \hline
            Stage 3 & Transformer~\cite{vaswani2017attention} &
            $
            \begin{bmatrix}
               & \text{embed dim }\TdimcRaw,~\text{head }\TdimcRaw/\TdimhRaw, &\\
               & \text{window size }\wsthreeRaw\times\wsthreeRaw, &\\
               & \text{mlp ratio }\mlpRaw &
            \end{bmatrix}\times\dthreeRaw $
            & $14\times14$ \\
            \hline
            Downsampling &
            \text{MBConv}~\cite{mobilenetv3}
            &
            $
            \begin{bmatrix}
                \text{embed dim } \TdimcRaw, \text{stride }2, \\
                \text{hidden/output dim } \TdimdRaw
            \end{bmatrix}\times 1
            $
            & $7\times7$ \\
            \hline
            Stage 4 & Transformer~\cite{vaswani2017attention} &
            $
            \begin{bmatrix}
               & \text{embed dim }\TdimdRaw,~\text{head }\TdimdRaw/\TdimhRaw, &\\
               & \text{window size }\wsfourRaw\times\wsfourRaw, &\\
               & \text{mlp ratio }\mlpRaw &
            \end{bmatrix}\times\dfourRaw $
            & $7\times7$ \\
            \hline
            Classifier & AvgPool+LayerNorm+Linear &
            \centering
            $
                \begin{bmatrix}
                    \text{output dim: the number of classes}
                \end{bmatrix}
            $
            & \\
            \Xhline{2\arrayrulewidth}
        \end{tabular}
    }
}
    \label{tab:arch}
    \vspace{-4mm}
\end{table*}

\vspace{-2mm}
\section{How fast the distillation is?}
\vspace{-1mm}
The proposed fast pretraining distillation is faster than the conventional distillation method by 29.8\% when using Florence model as the teacher (682M Params and 97.9 GFLOPs). More concretely, our method takes 92.4 GPU days to store the top-100 logits of Florence and 140.0 GPU days to pretrain TinyViT-21M (4.4 GFLOPs) with the saved logits for 90 epochs on ImageNet-21k, while the conventional distillation uses 330.9 GPU days due to limited batch size. Since the teacher logits per epoch are different and independent, they can be saved in parallel, instead of epoch-by-epoch in the conventional method. Besides, the saved logits can be reused for arbitrary student models, and avoid re-forwarding cost of the large teacher model.

\vspace{-2mm}
\section{Distillation with ground-truth}
\vspace{-1mm}
We compare the performance under the distillation with and without ground-truth. The student model is a variant of TinyViT-21M, equipped with talking head~\cite{talking_head} and shared blocks in Stage 4. As shown in Tab~\ref{tab:disgt}, the distillation with ground-truth would cause slight performance drops. This is probably because that not all the labels in ImageNet-21k~\cite{imagenet} are mutually exclusive. For example, it contains labels like ``chair'' and ``furniture'', ``horse'' and ``animal''~\cite{ridnik2021imagenet}, which are correlative pairs. Therefore, the one-hot ground-truth label could not describe an object precisely, and in some cases it suppresses either child classes or parent classes during training. By contrast, the soft labels generated by pretrained foundation models carry a lot of category relation information, that is helpful for distilling a small model, as presented in Fig. 3 of the main paper.


\begin{table*}[t]
\begin{center}
\caption{Comparison for pretraining distillation w/ and w/o ground truth (GT) labels. The student model is a variant of TinyViT-21M, pretrained for 90 epochs on ImageNet-21k and then finetuned for 30 epochs.
}
\label{tab:disgt}
\setlength{\tabcolsep}{8pt}{
\scalebox{1.0}{
\begin{tabular}{c|c|ccc}
\Xhline{2\arrayrulewidth}
IN-21k & Distillation & IN-1k & IN-Real & IN-V2 \\
Pretrained Teacher & Type & Top-1(\%) & Top-1(\%) & Top-1(\%) \\
\Xhline{2\arrayrulewidth}

\multirow{2}{*}{CLIP-ViT-L/14} & w/ GT & 84.3 & 88.5 & 73.6 \\
\cline{2-5}
 ~ & w/o GT & 84.5{\textcolor{YellowGreen}{\scriptsize{(+0.2)}}} & 88.8{\textcolor{YellowGreen}{\scriptsize{(+0.3)}}} & 74.4{\textcolor{YellowGreen}{\scriptsize{(+0.8)}}} \\
\hline
\multirow{2}{*}{Florence} & w/ GT & 84.2 & 88.5 & 73.7 \\
\cline{2-5}
~ & w/o GT & 84.9{\textcolor{YellowGreen}{\scriptsize{(+0.7)}}} & 89.0{\textcolor{YellowGreen}{\scriptsize{(+0.5)}}} & 74.9{\textcolor{YellowGreen}{\scriptsize{(+1.2)}}} \\
\Xhline{2\arrayrulewidth}
\end{tabular}
}
}
\end{center}
\vspace{-6mm}
\end{table*}


\end{document}